\documentclass{article}

\usepackage{microtype}
\usepackage{graphicx}
\usepackage{subfigure}
\usepackage{booktabs} % for professional tables

\usepackage{hyperref}
\usepackage{multicol}

\usepackage[accepted]{icml2024}

\usepackage{amsmath}
\usepackage{amssymb}
\usepackage{mathtools}
\usepackage{amsthm}
\usepackage{xspace}
\usepackage{amsmath,amsfonts,bm}

\newcommand{\pn}[1]{}

\newcommand{\Ms}{\mathcal{M}_S}
\newcommand{\Ml}{\mathcal{M}_L}

\newcommand{\red}[1]{{\color{red}#1}}

\newcommand{\Tandemce}{Tandem-CE\xspace}
\newcommand{\Tandemdistill}{Tandem-Distil\xspace}
\newcommand{\numsamples}{\textrm{num-samples}}
\newcommand{\gecko}{PaLM2-Gecko\xspace}
\newcommand{\xxxs}{PaLM2-XXXS\xspace}
\newcommand{\geckodistill}{PaLM2-Gecko-Distil\xspace}
\newcommand{\xxxsdistill}{PaLM2-XXXS-Distil\xspace}
\newcommand{\otter}{PaLM2-Otter\xspace}
\newcommand{\bison}{PaLM2-Bison\xspace}
\newcommand{\xtilde}[2]{\widetilde{x}_{#1}^{(#2)}}
\newcommand{\xss}[2]{{x}_{#1}^{(#2)}}
\newcommand{\ytilde}[2]{\widetilde{y}_{#1}^{(#2)}}
\newcommand{\yss}[2]{{y}_{#1}^{(#2)}}
\newcommand{\yhat}[2]{\widehat{y}_{#1}^{(#2)}}
\newcommand{\attn}[3]{\mathrm{Atn}^{(#3)}(#1 \vert #2)}
\newcommand{\ffn}[2]{\mathrm{FF}^{(#2)}(#1)}
\newcommand{\emb}[1]{\mathrm{Emb}(#1)}
\newcommand{\attns}[3]{\mathrm{Atn}_{\textrm{S}}^{(#3)}(#1 \vert #2)}
\newcommand{\ffns}[2]{\mathrm{FF}_{\textrm{S}}^{(#2)}(#1)}
\newcommand{\attnl}[3]{\mathrm{Atn}_{\textrm{L}}^{(#3)}(#1 \vert #2)}
\newcommand{\ffnl}[2]{\mathrm{FF}_{\textrm{L}}^{(#2)}(#1)}
\newcommand{\ffnTandem}[2]{\mathrm{FF}_{\textrm{Tandem}}^{(#2)}(#1)}
\newcommand{\ffndeepTandem}[1]{\mathrm{FF}_{\textrm{Tandem}}(#1)}
\newcommand{\embs}[1]{\mathrm{Emb}_{\textrm{S}}(#1)}
\newcommand{\embl}[1]{\mathrm{Emb}_{\textrm{L}}(#1)}
\newcommand{\xindex}[1]{x\left[#1\right]}

\def\ceil#1{\lceil #1 \rceil}
\def\floor#1{\lfloor #1 \rfloor}
\def\1{\bm{1}}

\DeclareMathAlphabet{\mathsfit}{\encodingdefault}{\sfdefault}{m}{sl}
\SetMathAlphabet{\mathsfit}{bold}{\encodingdefault}{\sfdefault}{bx}{n}

\usepackage{hyperref}
\usepackage{url}
\usepackage{array}
\usepackage{graphicx}
\usepackage{multirow}
\usepackage{enumitem}

\usepackage[capitalize,noabbrev]{cleveref}

\theoremstyle{plain}

\theoremstyle{definition}

\theoremstyle{remark}

\usepackage[textsize=tiny]{todonotes}
\newcommand{\pj}[1]{}

\icmltitlerunning{Tandem Transformers for Inference Efficient LLMs}

\begin{document}

\twocolumn[
\icmltitle{Tandem Transformers for Inference Efficient LLMs}

% It is OKAY to include author information, even for blind
% submissions: the style file will automatically remove it for you
% unless you've provided the [accepted] option to the icml2024
% package.

% List of affiliations: The first argument should be a (short)
% identifier you will use later to specify author affiliations
% Academic affiliations should list Department, University, City, Region, Country
% Industry affiliations should list Company, City, Region, Country

% You can specify symbols, otherwise they are numbered in order.
% Ideally, you should not use this facility. Affiliations will be numbered
% in order of appearance and this is the preferred way.
\icmlsetsymbol{equal}{*}

\begin{icmlauthorlist}
\icmlauthor{Aishwarya P S}{gdm}
\icmlauthor{Pranav Ajit Nair}{gdm}
\icmlauthor{Yashas Samaga}{gdm}
\icmlauthor{Toby Boyd}{gdm}
\icmlauthor{Sanjiv Kumar}{grnyc}\\
\icmlauthor{Prateek Jain}{equal,gdm}
\icmlauthor{Praneeth Netrapalli}{equal,gdm}
\end{icmlauthorlist}

% \icmlaffiliation{gri}{Google Research, India}
\icmlaffiliation{grnyc}{Google Research, New York City}
\icmlaffiliation{gdm}{Google DeepMind}

\icmlcorrespondingauthor{Aishwarya P S}{aishwaryaps@google.com}
\icmlcorrespondingauthor{Praneeth Netrapalli}{pnetrapalli@google.com}

% You may provide any keywords that you
% find helpful for describing your paper; these are used to populate
% the "keywords" metadata in the PDF but will not be shown in the document
\icmlkeywords{Machine Learning, ICML}

\vskip 0.3in
]

% this must go after the closing bracket ] following \twocolumn[ ...

% This command actually creates the footnote in the first column
% listing the affiliations and the copyright notice.
% The command takes one argument, which is text to display at the start of the footnote.
% The \icmlEqualContribution command is standard text for equal contribution.
% Remove it (just {}) if you do not need this facility.

%\printAffiliationsAndNotice{}  % leave blank if no need to mention equal contribution
\printAffiliationsAndNotice{\icmlEqualContribution} % otherwise use the standard text.

\begin{abstract}
The autoregressive nature of conventional large language models (LLMs) inherently limits inference speed, as tokens are generated sequentially.
% This creates a bottleneck despite the utilization of powerful neural accelerators (GPUs/TPUs).
While speculative~\cite{leviathan2023fast} and parallel~\cite{stern2018blockwise} decoding techniques attempt to mitigate this, they face limitations: either relying on less accurate smaller models for generation or failing to fully  leverage the base LLM's representations.

We introduce a novel architecture, Tandem Transformers, to address these issues. This architecture uniquely combines (1) a small autoregressive model and (2) a large model operating in block mode (processing multiple tokens simultaneously). The small model's predictive accuracy is substantially enhanced by granting it attention to the large model's richer representations.  On the PaLM2 pretraining dataset, a Tandem of \bison and \gecko demonstrates a 3.3\% improvement in next-token prediction accuracy over a standalone PaLM2-Gecko, offering a 1.16x speedup compared to a \otter model with comparable downstream performance. We further incorporate the Tandem model within the speculative decoding (SPEED) framework where the large model validates tokens from the small model. This ensures that the Tandem of \bison and \gecko achieves substantial speedup (around $1.14 \times$ faster than using vanilla \gecko in SPEED) while maintaining identical downstream task accuracy.

%It has been widely observed that the performance of large language models (LLMs) improves with scale~\cite{brown2020language}. However, serving very large models on a wide scale is infeasible due to the prohibitive cost of LLM inference. One of the key reasons for this high serving cost is the autoregressive nature of response generation, which leads to poor utilization of the underlying hardware accelerators such as GPUs/TPUs. In this work, we propose a novel architecture, called \emph{Tandem Transformers}, to mitigate the autoregressive component of LLMs.  By running the large model in block mode, we make more effective use of hardware accelerators, thereby reducing the overall latency. We evaluate Tandem Transformers in two settings: within speculative decoding (SPEED) framework~\cite{leviathan2023fast}, and as a stand alone model without SPEED and show overall latency improvements of up to $1.3$x in both of these settings.
\end{abstract}

\section{Introduction}

Despite significant advancements in inference optimization techniques \cite{leviathan2023fast,du2022glam,liu2023deja}, the widespread deployment of very large language models (LLMs) remains hindered by their substantial computational costs. A key factor contributing to high inference latency is the autoregressive generation process, where tokens are produced sequentially. This inherent limitation restricts the full utilization of ML accelerators (GPUs/TPUs), which are optimized for matrix-matrix multiplications rather than the matrix-vector operations prevalent in LLMs. Consequently, prompt processing (where all tokens are handled simultaneously) is significantly more efficient than autoregressive response generation.

On the other hand, it is not well understood how much capacity is required to understand the prompt/query/prefill (natural language understanding aka NLU) vs the capacity required to generate a response (natural language generation aka NLG). Current decoder-only LLM architectures tightly couple both these tasks. 

{\bf Tandem Transformers.} In this work, we investigate this fundamental question from an efficiency perspective. We propose Tandem Transformers, a novel architecture that allocates significantly more model capacity to prefill processing (NLU) compared to response generation (NLG). Our goal is to understand whether high-quality response generation can be maintained under this design.
Concretely,  Tandem Transformers consists of two models -- a small model $\Ms$ and a large model $\Ml$, where:
\begin{enumerate}[leftmargin=*,noitemsep,nolistsep]
    \item $\Ml$ processes the prompt/query.
    \item $\Ms$ generates the first $\gamma$ tokens (called a \emph{block}) autoregressively, while attending to the prompt/query representations generated by $\Ml$.
    \item $\Ml$ processes the $\gamma$ tokens generated by $\Ms$ together (i.e., in a non-autoregressive fashion) and computes their representations.
    \item $\Ms$ then generates the next $\gamma$ tokens autoregressively, while attending to representations of all tokens until the previous prefill \emph{block} generated by $\Ml$.
    \item This process is repeated until the response generation is complete.
\end{enumerate}
{\bf Tandem Transformer Training.} A projection layer is introduced to align the higher-dimensional representation space of $\Ml$ with that of $\Ms$. 
% A training strategy is designed to train $\Ml$, $\Ms$, and the projection layers based on this architecture.
For efficiency, we initialize $\Ml$ and $\Ms$ as independently trained, decoder-only models.

Experiments with Tandem (\bison, \gecko) (where \gecko $<$ \otter $<$ \bison, in terms of model size) demonstrate that the capacity needed for NLU vs NLG aspects of LLMs can indeed be decoupled, leading to a more efficient architecture without significant accuracy loss. Evaluation on benchmark datasets show that Tandem (\bison, \gecko) with block length $\gamma=3$ is substantially more accurate than \gecko, and comparable to \otter, while achieving approximately $1.16 \times$ lower inference latency than \otter. For example, on SuperGLUE~\cite{wang2019superglue}, the Tandem model is $3\%$ less accurate than \bison, $16\%$ more accurate than \gecko and $0.2\%$ less accurate than \otter, with $1.16 \times$ speedup over \otter.
 
{\bf Encoder-Decoder.} In contrast to an encoder-decoder architecture which would only process query/prefix through an encoder and then generate the entire response through a decoder, Tandem is able to generate only block-size $\gamma$ (say $=3$) tokens through the secondary model $\Ms$ and then refresh the entire prefill representations using primary model $\Ml$ which is critical to maintaining high accuracy. That is, by setting $\gamma=0$, Tandem can mimic decoder-only $\Ml$ model while setting $\gamma\rightarrow \infty$ leads to decoder-only $\Ms$ model. 

{\bf Tandem + SPEED.} For applications requiring output identical to the primary model, we propose Tandem + SPEED. The speculative decoding (SPEED) framework~\cite{leviathan2023fast} leverages the small model $\Ms$ in Tandem to  generate draft tokens, which are then verified by the large model $\Ml$. Crucially, the ability of $\Ms$ in Tandem to attend to $\Ml$'s representations significantly improves draft quality, reducing verification overhead compared to standard SPEED. For example, on the Reddit Posts dataset, using the $\Ms$ in Tandem as the drafter model in SPEED leads to about $11.24\%$ higher per-block acceptance rate compared to a vanilla secondary model. Finally, we show that Tandem Transformers can be further improved using logit distillation and their efficacy within SPEED can be improved using an adaptive block length parameter.

%While on most standard benchmarks, Tandem has significantly lower latency than the primary model, it's accuracy is still slightly subpar to that of primary model $\Ml$. This makes it challenging to compare it's efficacy compared to baseline models. To alleviate this challenge, we propose a Tandem+SPEED method which uses Tandem model in conjunction with the speculative decoding framework \cite{leviathan2023fast}. The key advantage of such a technique is that the output of Tandem+SPEED is {\em guaranteed} to match the output of baseline primary model $\Ml$, while still providing significantly lower latency than $\Ml$. At a high level, Tandem  generates $\gamma$ drafter tokens from the secondary model which can then be verified by primary model $\Ml$, and SPEED accepts the matching tokens and backtracks to a position where $\Ml$ and $\Ms$ deviates. Key advantage of Tandem over standard SPEED is that the drafter model (Tandem based $\Ms$) is significantly more accurate than a standard drafter model $\Ms$, leading to significantly higher number of accepted tokens and smaller amount of backtracking. 

%{\bf Adaptive Gamma.} 

{\bf Contrast with Parallel Decoding and Distillation.} Recently multiple speculative or parallel decoding style techniques have been proposed in the literature \cite{leviathan2023fast,kim2023speculative,stern2018blockwise}. These techniques attempt to generate a draft of tokens using a relatively inexpensive drafter model. Parallel decoding attempts to generate multiple drafter tokens in parallel by learning classifiers on top of output of primary model $\Ml$ while speculative decoding could provide significantly better drafts by using a small, but auto regressive model. In contrast, Tandem is a {\em stand alone} model on its own and doesn't natively require verification by $\Ml$ to generate reasonable outputs (see benchmark numbers in Table~\ref{tab:flat-downstream}). Furthermore, Tandem + SPEED is able to use representations of $\Ml$ while still generating tokens autoregressively, which is able to provide overall much better tradeoff in terms of token quality vs model latency for the drafter. Finally, recent works have also shown the efficacy of logit distillation for training better drafter models within SPEED~\cite{zhou2023distillspec}. Our approach is complementary, and can be combined with distillation.

{\bf Empirical Results for Tandem + SPEED.} Finally, we conduct extensive latency evaluation on TPUv5e for both standa alone and SPEED versions of Tandem (\bison, \gecko) with \bison and \gecko being the primary $\Ml$ and secondary $\Ms$ model, respectively. In particular, on multiple datasets, we observe that Tandem + SPEED with distillation can be at least $2.19 \times$ faster than the baseline \bison model while ensuring same output quality. Furthermore, compared to standard SPEED with $\Ms$ being secondary model, our model is $1.11\times$ to $1.17\times$ faster. An adaptive block length in SPEED further helps reduce Tandem's latency by $1.04\times$ to $1.09\times$ on multiple datasets. Finally, we demonstrate that our results also hold for practical settings like batch-size $>1$. 

{\bf Contributions.} In summary, following are the key contributions of the work: 
\begin{enumerate}[leftmargin=*,noitemsep,nolistsep]
    \item Tandem architecture: A novel architecture to disaggregate prompt/prefill processing capacity from response generation. 
    \item Tandem + SPEED: Improved speculative decoding leveraging Tandem's superior drafting for guaranteed output equivalence with lower latency.
    \item Adaptive Block Length: Enhances Tandem + SPEED by dynamically adjusting drafted token count.
    \item TPUv5e evaluation: End-to-end evaluation on TPUv5e with \bison being the primary model. A distilled Tandem + SPEED is $2.4 \times$ faster compared to vanilla \bison model and $1.11-1.17 \times$ faster compared to distilled $\Ms$ + SPEED \cite{leviathan2023fast} applied in the same setting. 
\end{enumerate}

{\bf Outline of the Paper.} The rest of the paper is organized as follows. We briefly review related work in Section~\ref{sec:related}. In Section~\ref{sec:idea}, we present the main ideas and the design of Tandem Transformers architecture. Section~\ref{sec:exp} presents the experimental results on Tandem Transformers. We then conclude with some future directions in Section~\ref{sec:conc}.

% {\color{red} TODO:
% \begin{itemize}
%     \item Picture in terms of blocks
%     \item Formally define a Tandem architecture in equations. (output computation)
%     \item Large model vs small model, MHA vs MQA etc.
%     \item subsection on training. three training settings. 1. large model frozen 2. both models training but loss on small model only 3. loss on both models.
%     \item subsection on inference, recomputation of large model representations. some differences between frozen and non-frozen in terms of using the free token. Updating KV cache.
%     \item SPEED + Tandem
% \end{itemize}

% }

% \input{intro-deepTandem}
\section{Related Work}\label{sec:related}
{\bf Encoder-Decoder models.} Encoder-decoder Transformer architectures are widely used for specific tasks such as machine translation~\cite{vaswani2017attention}. Given the computational inefficiency of autoregressive decoding, several works have explored using a large encoder with a small decoder. Our work can be seen as extending these ideas to use an encoder-decoder model for the decoder itself.

\vspace{2.5mm}
{\bf Mixture of Experts (MoE)/Sparsity based Approaches.} Mixture of experts~\cite{du2022glam} and sparsity based approaches~\cite{li2022lazy} have also been studied for optimizing inference cost of LLMs. However these approaches are complementary to the approaches proposed in our paper. For example, either or both the large model $\Ml$ and small model $\Ms$ can be an MoE or sparse model.

\vspace{2.5mm}
{\bf Distillation.} Since the seminal paper~\cite{hinton2015distilling}, distilling the knowledge of a large model to a smaller model by using the logits of large model as a training target has been widely used in several settings. Our work can be seen as a more general version of distillation for Transformers, where the small model can directly refer to large model representations for tokens from previous blocks. Furthermore, our experiments (see Section~\ref{sec:exp}) show that our techniques are complementary to logit distillation, and provide additional gains on top of vanilla logit distillation.

\vspace{2mm}
{\bf Speculative Decoding (SPEED).} Speculative decoding~\cite{leviathan2023fast,kim2023speculative} is a framework to reduce inference latency of LLMs without affecting their quality, which has shown substantial improvements in LLM inference. We demonstrate that Tandem Transformers can be used within the SPEED framework, improving the efficacy of SPEED. Multiple drafters have been explored in the context of SPEED,  Speculative Decoding~\cite{leviathan2023fast} uses a standalone drafter, REST~\cite{he2023rest} use a retrieval based drafter, DistilSpec~\cite{zhou2023distillspec} uses a distillation based drafter, MEDUSA~\cite{medusa} uses $k$ MLP heads on top of the primary model's output representations to predict the next $k$ tokens in parallel for drafting. As of now distillation based drafters seem to perform the best. As we demonstrate in Section~\ref{sec:exp}, Tandem is able to provide a significantly more powerful drafter, thus providing better draft of tokens leading to lower latency. 
\section{Tandem Transformers}\label{sec:idea}
In this section, we will describe Tandem Transformers architecture, it's training and inference.  %two Transformer models $\Ms$ and $\Ml$ (small and primary model respectively).
\begin{figure*}
    \centering
    \includegraphics[scale=0.5]{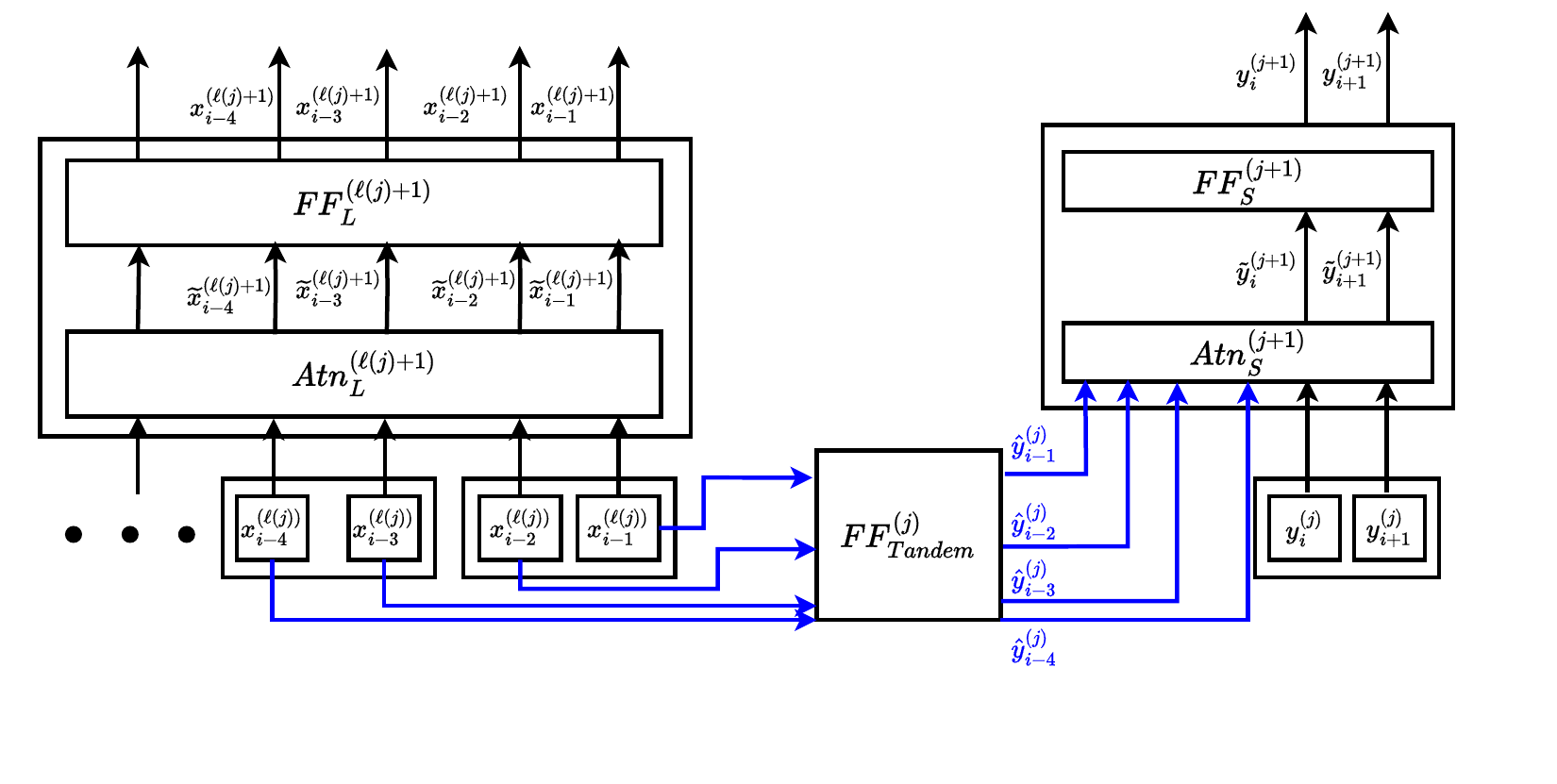}
    \vspace{-12mm}
    \caption{Training of Tandem Transformers with a block length $\gamma = 2$. $\mathrm{Atn}_L^{(\ell(j)+1)}$ and $\mathrm{FF}_L^{(\ell(j)+1)}$ denote the attention and feedforward blocks in the $(\ell(j)+1)^\textrm{th}$ layer of $\Ml$, while $\mathrm{Atn}_L^{(j+1)}$ and $\mathrm{FF}_L^{(j+1)}$ denote those of $(j+1)^\textrm{th}$ layer of $\Ms$. $\Ml$ processes the tokens as a standard decoder Transformer. $\Ms$ on the other hand processes the 
    tokens in the $\left(\frac{i}{\gamma}\right)^\textrm{th}$ block using its own representations $\yss{i}{j}$ and $\yss{i+1}{j}$, but while attending to the representations of all tokens from the previous block from the $(\ell(j)+1)^\textrm{th}$ layer of $\Ml$ passed through a feedforward layer $\mathrm{FF}_{\textrm{Tandem}}^{(j)}$.}
    \label{fig:training}
\end{figure*}

{\bf Standard (Decoder) Transformer.} Given a sequence $t_1, t_2, \cdots, t_S$ of $S$ tokens as inputs, where $t_i$ corresponds to the $i^\textrm{th}$ token id, a standard decoder Transformer with $L$ layers executes as follows:
\begin{align}
    \xtilde{i}{j+1} &= \attn{\xss{i}{j}}{\xss{\leq i}{j}}{j+1} \nonumber \\
    \xss{i}{j+1} &= \ffn{\xtilde{i}{j+1}}{j+1} \qquad \mbox{ for } j = 0,\cdots,L-1, \label{eqn:decoder}
\end{align}
where $\xss{i}{0} = \emb{t_i}$ is the embedding of $t_i$, $\xss{i}{j}$ is the representation after the $j^\textrm{th}$ layer and $\attn{\cdot}{\cdot}{j}$ and $\ffn{\cdot}{j}$ are the $j^\textrm{th}$ attention and feedforward layers respectively~\cite{vaswani2017attention}. Note that the attention is purely causal (i.e., the $i^\textrm{th}$ token attends only tokens $t_{k}$ for $k\leq i$) since we are considering a decoder-only Transformer.

{\bf Tandem Transformer.} A Tandem Transformer model comprises of a primary model $\Ml$ and a secondary model $\Ms$. Typically, $\textsc{sizeof}(\Ml)\gg \textsc{sizeof}(\Ms)$.  Given a sequence of tokens $t_1, t_2, \cdots, t_S$ as inputs, the primary model $\Ml$ processes these tokens just like a standard (decoder) Transformer  \eqref{eqn:decoder}. 

Let $\gamma$ be the block length parameter, and $L_S$ and $L_L$ be the number of layers of the secondary model and primary model, respectively. Let  $\ell: [L_S] \rightarrow [L_L]$ be a layer assignment function from secondary model to primary model. The secondary model attends to the primary model's representations for all tokens from the previous blocks. More formally, we have:
\begin{align}
    \yhat{i}{j} &= \ffnTandem{\xss{i}{\ell(j)}}{j} \nonumber\\
    \ytilde{i}{j+1} &= \attns{\yss{i}{j}}{\yhat{\leq k}{j}, \yss{\left[ k+1, i \right]}{j}}{j+1} \mbox{ where } k = \floor{\frac{i}{\gamma}}*\gamma \nonumber \\
    \yss{i}{j+1} &= \ffns{\ytilde{i}{j+1}}{j+1} \qquad \mbox{ for } j = 0,\cdots,L_S-1, \label{eqn:Tandem}
\end{align}
where $\xss{i}{j}$ and $\yss{i}{j}$ denote the $j^{\textrm{th}}$ layer representation of the $i^\textrm{th}$ token under $\Ml$ and $\Ms$ respectively, $\ffnTandem{\cdot}{j}$ denotes a feedforward layer that converts the representation $\xss{i}{\ell(j)}$ of the $i^\textrm{th}$ token from the $\ell(j)^\textrm{th}$ layer of the primary model, to a representation $\yhat{i}{j}$ of the same $i^{\textrm{th}}$ token for the $j^\textrm{th}$ layer of the secondary model, and $\attns{\cdot}{\cdot}{j}$ and $\ffns{\cdot}{j}$ denote the attention and feedforward blocks respectively in the $j^{\textrm{th}}$ layer of the secondary model $\Ms$. The final output of the Tandem model is $\yss{}{L_S}$. We note that the primary and the secondary model can vary in almost all scale parameters such as representation dimensions, expansion factors of feedforward layers, number of attention heads, etc. as well as whether the attention is multi-head or multi-query, etc. In all of our experiments, we take $\ffnTandem{j}{\cdot}$ to be linear projection layers.

{\bf Training.}
Given a block length parameter $\gamma$, we partition the training sequence into blocks, each consisting of $\gamma$ consecutive tokens. Consider the autoregressive prediction of the $j^\textrm{th}$
  token (for some $j \leq \gamma$) within the $i^\textrm{th}$ block. The input to the secondary model $\Ms$ is the previous token. Crucially, within the attention blocks of $\Ms$: 
  % \begin{itemize}
  \begin{itemize}[leftmargin=*,noitemsep,nolistsep]
      \item Key/value pairs for all tokens up to the $j^\textrm{th}$ token in the \emph{current block} are computed by $\Ms$ itself. 
      \item Key/value pairs for tokens in previous blocks are computed by the primary model $\Ml$. A projection/Tandem feedforward layer then aligns the representational dimensions from $\Ml$ to $\Ms$, as described in Equation~\eqref{eqn:Tandem}.
  \end{itemize}
  
We explore multiple training configurations for Tandem Transformers:
\begin{itemize}[leftmargin=*,noitemsep,nolistsep]
    \item {\bf Primary Model Frozen.}  Only the secondary model parameters $\Ms$ and the Tandem feedforward layer $\mathrm{FF}_S^{(j)}$ are updated. Loss is applied solely to the secondary model's output $\yss{}{L_S}$ (Equation~\eqref{eqn:Tandem}).
    \item {\bf Both Models Trained, Loss on Secondary Outputs.}  Similar to the above, loss is applied to the secondary model's output. However, both $\Ml$ and $\Ms$, along with $\mathrm{FF}_S^{(j)}$ are trained.
    \item {\bf Both Models Trained, Loss on Both Outputs.}  The combined loss incorporates both the primary model's outputs $\xss{}{L_L}$ and the secondary model's outputs $\yss{}{L_S}$.
\end{itemize}
For training efficiency, we initialize the primary and secondary models with high quality pretrained checkpoints, and then continue pretraining the Tandem architecture for a small number of additional steps. In particular, we use the pretrained \bison and \gecko checkpoints to initialize $\Ml$ and $\Ms$ respectively. In this setting, we found that {\bf Primary Model Frozen} approach provides the best accuracy. Our \Tandemce model is obtained by using cross entropy (CE) loss on the output of the secondary model as described above.

\Tandemdistill: To further enhance $\Ms$'s quality, we apply a distillation loss on its predictions, using the logits of the pretrained $\Ml$ as targets with CE loss. This aligns naturally with the Tandem architecture, as $\Ms$ already incorporates representations from $\Ml$. 

The \Tandemdistill model follows a two stage training setup, where initially it is trained to minimize the CE loss with respect to the ground truth labels, and in the second stage a weighing factor of $\lambda=0.5$ is used to balance the CE loss with respect to ground truth labels and the CE logit distillation loss with respect to the outputs of the \bison model. We note that \Tandemdistill in general performs better than \Tandemce.

\begin{figure*}[t]
\centering
% \begin{minipage}{0.45\textwidth}
%     \centering
    \includegraphics[width=1\textwidth]{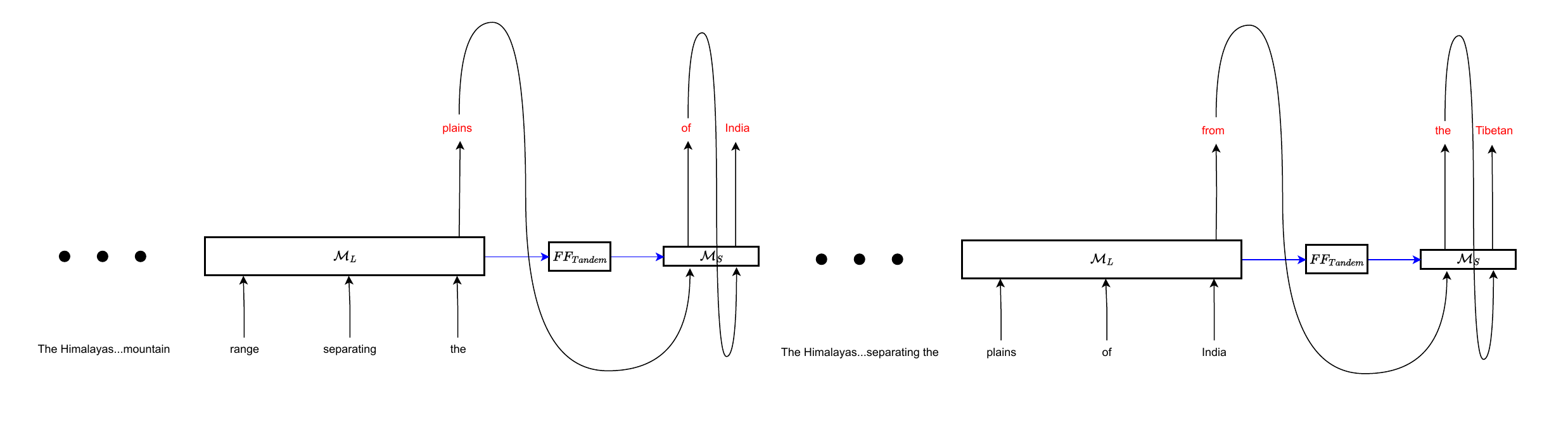}
% \end{minipage}
% \begin{minipage}{0.45\textwidth}
%     \centering
%     \includegraphics[width=1\textwidth]{figures/TandemInference2_freetoken.drawio.pdf}
% \end{minipage}
\vspace{-12mm}
    \caption{Inference of Tandem Transformers \emph{with} free token from the primary model $\Ml$. (left) First block prediction. (right) Second block prediction. Given the query \emph{The Himalayas are a mountain range separating the}, $\Ml$ first processes this query and produces the first response token \red{plains}. When we use this prediction from $\Ml$, this is directly fed as an input to the secondary model $\Ms$, which autoregressively produces \red{of India} for the first block with $\gamma = 2$. In the second block, the entire response from the first block \red{plains of India} is fed to the primary model $\Ml$, which again produces the next response token \red{from}, and then the secondary model $\Ms$ produces the next two tokens of the block \red{the Tibetan} autoregressively. The eventual output of the model will be \red{plains of India from the Tibetan ...}.}
    \label{fig:inference-free-token}
\end{figure*}

\paragraph{Inference.} The inference process begins with the primary model ($\Ml$) processing the prompt and generating representations for all prompt tokens.  The secondary model ($\Ms$) then autoregressively generates the first block of $\gamma$ response tokens. Crucially, $\Ms$ attends to the primary model's representations, aligned via the projection layer. 

Once the first response block is generated, the primary model ($\Ml$) processes these tokens and computes their representations. We consider two inference configurations:
\begin{itemize}[leftmargin=*,noitemsep,nolistsep]
\item {\bf Representation Generation + Token Prediction (Figure~\ref{fig:inference-free-token})}. $\Ml$ additionally predicts the next token.
\item {\bf Representation Generation Only (Appendix~\ref{no-token}, Figure~\ref{fig:inference-no-free-token})}. $\Ml$ solely generates representations for the response block.
\end{itemize}
In both configurations, the representations generated by $\Ml$ are used by the secondary model ($\Ms$) to generate the subsequent block of $\gamma$ response tokens.  Also note that, as in training, $\Ms$ attends to its own representations for all previous tokens within the current block.

To disaggregate query and response generation, we use {\bf Representation Generation Only} for processing the input query/prefix. However, for subsequent blocks where the prefill (query+generated response till this point) is processed, we use {\bf Representation Generation + Token Prediction} from $\Ml$.

% Key Points

% Representational Alignment: The projection layer ensures compatibility between the representational spaces of $\Ml$ and $\Ms$.
% Flexibility: The two inference configurations allow for exploring the trade-off between the primary model's computational cost and its potential improvement in overall accuracy.

% \paragraph{Inference}: During inference, the prompt is processed by the primary model $\Ml$, which generates representations for all the prompt tokens. The secondary model $\Ms$, then generates the first block of $\gamma$ response tokens autoregressively by attending to the representations of the primary model, passed through the projection layer. Once the first block of response tokens are generated, the primary model $\Ml$ processes these tokens and generates representations for this block of tokens. 

% Here, we can either limit the role of primary model to generating the representations for these tokens (as depicted in Figure~\ref{fig:inference-no-free-token}, or use it to generate the next token as well (as depicted in Figure~\ref{fig:inference-free-token}. In either case, the representations generated by $\Ml$ are then used by the secondary model $\Ms$ for generating the second block of $\gamma$ response tokens. Similar to training, the secondary model $\Ms$ attends to its own representations for all previous tokens in the current block.

Depending on the training protocol -- specifically, whether primary model outputs are reliable -- we may optionally allow the primary model ($\Ml$) to generate the first token of the subsequent block (processing $\gamma+1$ tokens).  Crucially, in this scenario, we must ensure the following: the keys and values associated with the next block's first token, computed by $\Ml$, are not overwritten when the secondary model ($\Ms$) executes its attention layers.

{\bf Inference-Time Block Length Flexibility.} 
While we train Tandem Transformers with a fixed block length $\gamma$, the architecture supports arbitrary $\gamma$ values during inference. Larger $\gamma$ values generally improve efficiency by maximizing the primary model's ($\Ml$) utilization of accelerator hardware. Although Tandem is trained with a fixed $\gamma$, in SPEED evaluations we find that the optimal $\gamma$ is often much larger, indicating the robustness of Tandem to changes in $\gamma$ at inference time.

% Depending on the training protocol (in particular whether primary model outputs are reliable or not), we can also let the primary model $\Ml$ generate the first token of the next block  (i.e., it will process $\gamma+1$ tokens instead of $\gamma$ tokens). In this case, it is important to ensure that the keys and values, of this first token of the next block, in the attention layers of the secondary model, computed during the execution of the primary model $\Ml$ are not overwritten during the execution of the secondary model $\Ms$.
% \paragraph{Different block length at inference time}: While we train the Tandem model with a specific block length $\gamma$, it can be used during inference time with any other $\gamma$ as well. Higher $\gamma$ lets us execute the primary model $\Ml$ using the accelerator hardware more effectively, and will be more efficient. Empirically, we observe that the quality of Tandem model is robust to changes of block length $\gamma$ during inference time.
% \pn{Give numbers in a table.}

\subsection{Tandem + SPEED: Tandem in the Speculative Decoding Framework}
SPEED mitigates the inefficiency of autoregressive generation using a smaller drafter/secondary model to generate tokens and a larger verifier/primary model to confirm them. SPEED guarantees output quality matching the verifier, but its efficacy hinges on the drafter's ability to generate long, accurate draft sequences. Tandem Transformers are uniquely suited for this framework, with our secondary model $\Ms$ acting as the ``drafter" and  primary model $\Ml$ acting as the ``verifier".

% So, our \Tandemce+ SPEED method uses \Tandemce method to train $\Ms$ and projection layers and \Tandemdistill+ SPEED uses \Tandemdistill to train the same. Similar to standard Tandem, we use $\Ml$ to generate {\bf representations only} during the query/prefix processing and use {\bf representations and token} during the subsequent blocks. During inference, $\Ms$  generates $\gamma$ logit outputs: $[y^S_1, \dots, y^S_\gamma]$; inference time $\gamma$ can be different than $\gamma$ used during training. We then process the prefill and additional $\gamma$ tokens generated by $\Ms$, to generate the representations as well $\Ml$'s logit outputs: $[y^P_1, \dots, y^P_\gamma]$. Similar to SPEED, we ``accept" first $\alpha$ tokens till $\Ml$ and $\Ms$ agree on the output logits. That is, 
%$$\alpha=min\ i\ s.t.\ Dist(y^S_i, y^P_i)>\tau,$$
%where $Dist(y^S_i, y^P_i)$ is the total-variation distance \cite{} between union of top-k elements of $y^S_i$ and $y^P_i$ and $\tau$ is a fixed threshold.  Hence, the first $\alpha$ set of tokens are sampled from $[y^P_1, \dots, y^P_\alpha]$ using the prescribed sampling temperature, and we then again start response generation from that step; note that $\Ml$ is giving out the $\alpha$-th token instead of the $\Ms$ whose output might be very different for $\alpha$-th token. 

Given a Tandem model, we use $\Ml$ to process the query/prefix and generate representations for them. $\Ms$ uses these and produces a draft for the first $\gamma$ tokens autoregressively. $\Ml$ then verifies this entire block simultaneously and identifies the first location $i$ where the draft token is deemed incorrect by $\Ml$ ($i=\gamma+1$, if all the draft tokens are verified successfully). We take the output of the large model for the $i^{\textrm{th}}$ token, and the small model $\Ms$ then continues to generate draft tokens from the $(i+1)^{\textrm{th}}$ position onwards, while using the representations of \emph{all the previous tokens} from the large model $\Ml$. This process continues until a full response is generated.

% \pj{NEES TO BE FIXED}
The above process can be generalized to the setting, where we generate multiple full responses for the same query, we refer to it as \numsamples, for example to eventually rank these responses and select the ``best" response~\cite{DBLP:journals/corr/abs-2310-17022}. In this case, the location of the rejected token can vary across the different samples being generated.
% That is, for every sample $j$ in the batch of samples, we first compute $\alpha_j$ using the output logits for each sample from $\Ml$ and $\Ms$. We then set the overall {\em batch-level} $\alpha$ to be $\alpha=\min_j \alpha_j$, and accept the first $\alpha$ tokens for each sample and start further token generation from that point on. 

Similarly, the above approach generalizes to larger batch sizes as well, when we are simultaneously processing multiple queries together.
Practical systems potentially use both \numsamples \text{ } and batch-size to be $>1$ but latency gains for Tandem + SPEED depend on overall batch-size which is $\numsamples\times batch\_size$. So, for simplicity we focus only on $\numsamples>1$ and fix batch-size to be $1$\footnote{Note that it is more challenging to obtain latency improvements with increasing $\numsamples$, compared to that in batch size since, even without any of these optimizations such as SPEED etc., larger \numsamples obtain better efficiency on all layers while larger batch size obtains better efficiency only on feedforward and softmax layers, and not the attention layer.}. 
%which are generated by both $\Ml$ and $\Ms$ while backtrack our inference to the token where there is first disagreement between $\Ml$ and $\Ms$. 

\begin{table}[t!]
\centering
\caption{Accuracy and cross entropy (CE) loss of Tandem Transformers with respect to ground truth labels as well as the predictions of the primary model $\Ml$, \bison. As is clear from the results, the Tandem model of \gecko and \bison substantially outperforms the stand alone \gecko model.}
\vspace{2mm}
\resizebox{\columnwidth}{!}{
\begin{tabular}{lcccc}\toprule
 & \begin{tabular}[c]{@{}c@{}}PaLM2-\\Gecko\end{tabular} & \begin{tabular}[c]{@{}c@{}}PaLM2-\\Gecko-Distil\end{tabular}  & \begin{tabular}[c]{@{}c@{}}Tandem-CE\\\textbf{(ours)}\end{tabular}  & \begin{tabular}[c]{@{}c@{}}Tandem-\\Distil \textbf{(ours)}\end{tabular} \\
\midrule
 Accuracy (GT) & $55.06$ & $56.50$ & $58.35$ & $\mathbf{58.61}$ \\ 
 CE loss (GT) & ~$2.14$ & ~$2.12$ & $~\mathbf{1.94}$ & ~$1.99$ \\
 Relative accuracy & $74.64$ & $75.30$ & $80.00$ & $\mathbf{81.00}$ \\ 
 % Relative TV distance & $0.391$ & $0.318$ & $0.1775$ & $\mathbf{0.141}$ \\ [1ex]
Relative TV distance & $0.391$ & $0.318$ & $0.178$ & $\mathbf{0.141}$ \\
\bottomrule
\end{tabular}
}
\label{tab:flat-pretraining}
\end{table}

{\bf Adaptive Block Length.} While standard SPEED uses a fixed block length $\gamma$, we introduce an adaptive approach. We train a relatively small $2$-layer multi-layer perceptron -- {\em router MLP} --  model to predict whether the current draft token from $\Ms$ is likely to be accepted by the primary model $\Ml$. At each timestep, we compare the prediction of this small model to a threshold $\tau$, deciding whether to: a. Verify with $\Ml$, or b. Continue drafting with $\Ms$. 

Input features to the router MLP are: $\Ms$'s entropy over the current token's vocabulary distribution, top-$k$ probabilities for the current token for an appropriate $k$, and $\Ms$'s model embeddings corresponding to these top-$k$ most probable tokens. We train the router MLP to predict the probability of disagreement using cross-entropy loss, with ground truth being: $TV(y^S_j,y^P_j)$, where $TV(y^S_j,y^P_j)$ is the total variation (TV) distance between the output logits of $\Ms$ and $\Ml$ for the $j^\textrm{th}$ token. %have as the label  to be a very strong indicator of whether the two would agree, and hence, use it as the target to train the MLP for binary classification.

\section{Experiments}\label{sec:exp}
In this section, we present experimental results evaluating Tandem Transformer models. Except for the new architecture of Tandem Transformers, we generally follow the same training protocols as described in~\cite{palm2}, including the training dataset, optimizer, etc.

%\subsection{Training Details}
%We train only the secondary model $\Ms$ and the projection/Tandem feedforward layers in the Tandem setup, while freezing the primary model $\Ml$. 
\vspace{2.5mm}
{\bf Further Training Details.} For both \Tandemce and \Tandemdistill, we initialize the secondary model $\Ms$ to be the pretrained \gecko, while freezing primary model $\Ml$ to be the pretrained \bison~\cite{palm2}. The projection/Tandem feedforward layers are chosen to be linear layers and initialized randomly. Both the Tandem models -- \Tandemce and \Tandemdistill -- are trained with a block length of $\gamma = 2$. 
For our evaluation within the SPEED framework, we consider a logit distillation version of \gecko, called \geckodistill, which is initialized with the \gecko model and then trained using logit distillation, similar to the second phase of training of the \Tandemdistill model, since distillation has been shown to help improve the secondary models in SPEED~\cite{zhou2023distillspec}.

%{\bf Tandem+SPEED hyperparameters.} We use $\tau=0.01$ as threshold to accept tokens generated from $\Ms$ (see Section~\ref{sec:Tandemspeed}). 

\vspace{2mm}
{\bf Adaptive Block Length in SPEED.} 
We train a small, $2$-layer MLP model to predict whether the current drafter token from $\Ms$ is likely to be accepted by primary model $\Ml$. We set $\tau=0.8$ as the threshold to determine if $\Ms$ can continue generating more tokens. 
% To calculate the entropy and top-$k$ probabilities, the drafter model $\Ms$'s temperature is set to $\tau = 1$ and the primary model's temperature is set to $\tau = 0.4$.

\begin{table*}[t!]
\caption{End-to-end latency gain of various secondary models, when used within the SPEED framework with \bison as the primary model. The secondary models we consider are: \geckodistill and Tandem-Distil. Since \Tandemdistill has better acceptance rate compared to \geckodistill, e.g., for $\gamma = 5$, \Tandemdistill has, on average, $11.24\%$ more tokens accepted compared to \geckodistill, for each secondary model, and on each dataset, we use the optimal block length $\gamma$ parameter. We consider two settings, one where we generate a single response and another where we generate $4$ responses for the given query. The third and fourth column provide the speedup by using \geckodistill and Tandem models respectively, with respect to the \bison model. The last column indicates the relative gain of using the Tandem model as the secondary model in SPEED, instead of \geckodistill. The results clearly demonstrate the additional improvements Tandem obtains, on top of logit distillation.}
\vspace{2mm}
% \begin{center}
\centering
% \resizebox{\textwidth}{!}{
\begin{tabular}{lcccc}
 \toprule
 \multirow{2}{*}{Dataset} & num- & PaLM2-Gecko-Distil & Tandem-Distil  & Tandem-Distil \\ & samples & (baseline) & (\textbf{ours}) & (\textbf{ours}; relative gain) \\ 
 \midrule
 \multirow{2}{*}{\centering Reddit} & $1$&	$2.169 \times$ \text{   
 }($\gamma=7$) &	$2.471 \times$ \text{   
 }($\gamma=7$) & $\mathbf{1.139} \times$ \\ 
 % \cline{2-5}
 & $4$& $ 1.919 \times$ \text{ 
 }($\gamma=5$)& $2.234\times$ \text{   
 }($\gamma=7$)& $\mathbf{1.164} \times$	\\ 
 \midrule
\multirow{2}{*}{\centering CNN/DailyMail} & $1$&	$ 2.219 \times$ \text{ 
 }($\gamma=7$)&	$2.473 \times$ \text{   
 }($\gamma=7$)& $\mathbf{1.115} \times$ \\
 % \cline{2-5}
 & $4$& $1.940 \times$ \text{ }($\gamma=5$)& $2.190 \times$ \text{   
 }($\gamma=7$)& $\mathbf{1.129} \times$\\
 \midrule
\multirow{2}{*}{\centering LM1B} & $1$& $2.348 \times$ \text{   
 }($\gamma=7$)&	$2.610 \times$ \text{   
 }($\gamma=7$)& $\mathbf{1.112} \times$  \\
 % \cline{2-5}
 & $4$& $ 2.011 \times$ \text{  
 }($\gamma=5$)& $2.359 \times$ \text{   
 }($\gamma=7$)& $\mathbf{1.173} \times$	\\
 \bottomrule
\end{tabular}
% }
% \end{center}
\label{tab:flat-latency}
\end{table*}

\begin{table*}[t!]
\caption{Standalone evaluation of the Tandem model. The first five rows present downstream evaluations of the Tandem Transformers on a variety of generative and ranking tasks. We see that the Tandem model substantially improves upon the performance of stand alone \gecko model, and is on par with the \otter model. On the other hand, the latency evaluations in the last row demonstrate that the Tandem model is about $1.16 \times$ faster than the \otter model.}
\vspace{2mm}
% \begin{center}
\centering
% \resizebox{\columnwidth}{!}{
\begin{tabular}{lccccc}
 \toprule
 \multirow{2}{*}{Dataset} & \gecko & Tandem-CE & Tandem-Distil & \otter & \bison \\ & & \textbf{(ours)} & \textbf{(ours)} & & \\ 
 \midrule
 Generative-tasks & $28.8$ & $37.1$ & $44.0$ & $51.1$ & $57.5$\\
 % \hline
 % GPT3-Rank & 57.1 & 70.3 & 70.2 & & 73.6\\ [1ex]
 % \midrule
 MBPP & $4.8$ & $13.8$ & $21.2$ & $20.8$ & $30.4$ \\ 
 % \midrule
 WMT22-1shot-to-nonenglish & $35.1$ & $37.4$ & $44.1$ & $48.4$ & $50.5$ \\ 
 % \midrule
 TydiQA-GoldP & $55.0$ & $65.7$ & $69.0$ & $69.7$ & $73.4$ \\ 
 % \hline
 % \midrule
 SuperGLUE & $62.8$ & $78.5$ & $78.8$ & $79.0$ & $81.5$ \\
 Speedup over \bison & $6.40 \times$ & $2.75 \times$ & $2.75 \times$ & $2.36 \times$ & $1 \times$ \\ 
 \bottomrule
\end{tabular}
% }
% \end{center}
\label{tab:flat-downstream}
\end{table*}

% \begin{table*}[t!]
% \caption{Standalone evaluation of the Tandem model. The first five rows present downstream evaluations of the Tandem Transformers on a variety of generative and ranking tasks. We see that the Tandem model substantially improves upon the performance of stand alone \gecko model, and is on par with the \otter model. On the other hand, the latency evaluations in the last row demonstrate that the Tandem model is about $1.16 \times$ faster than the \otter model.}
% \vspace{2mm}
% % \begin{center}
% \centering
% % \resizebox{\columnwidth}{!}{
% \begin{tabular}{lccccc}
%  \toprule
%  \multirow{2}{*}{Dataset} & \gecko & Tandem-Distil & \otter & \bison \\ & & \textbf{(ours)} & & \\ 
%  \midrule
%  Generative-tasks & $28.8$ & $44.0$ & $51.1$ & $57.5$\\
%  % \hline
%  % GPT3-Rank & 57.1 & 70.3 & 70.2 & & 73.6\\ [1ex]
%  % \midrule
%  MBPP & $4.8$ & $21.2$ & $20.8$ & $30.4$ \\ 
%  % \midrule
%  WMT22-1shot-to-nonenglish & $35.1$ & $44.1$ & $48.4$ & $50.5$ \\ 
%  % \midrule
%  TydiQA-GoldP & $55.0$ & $69.0$ & $69.7$ & $73.4$ \\ 
%  % \hline
%  % \midrule
%  SuperGLUE & $62.8$ & $78.8$ & $79.0$ & $81.5$ \\
%  Speedup over \bison & $6.40 \times$ & $2.75 \times$ & $2.36 \times$ & $1 \times$ \\ 
%  \bottomrule
% \end{tabular}
% % }
% % \end{center}
% \label{tab:flat-downstream}
% \end{table*}

\vspace{2.5mm}
\subsection{Performance Evaluation}
We compare the performance of \Tandemce and \Tandemdistill against \gecko, \geckodistill, \otter and \bison on several downstream tasks.

For downstream task evaluation, we compare on SuperGLUE \cite{wang2019superglue}, TydiQA \cite{Clark2020TyDiQA}, a large collection of generation tasks, which we call Gen-tasks (comprising of SQuADv2 \cite{DBLP:conf/acl/RajpurkarJL18}, Natural Questions \cite{DBLP:journals/tacl/KwiatkowskiPRCP19}, TriviaQA \cite{DBLP:conf/acl/JoshiCWZ17}, WebQuestions \cite{DBLP:conf/emnlp/BerantCFL13} and Lambada \cite{DBLP:conf/acl/PapernoKLPBPBBF16}), MBPP \cite{Austin2021ProgramSW}, and WMT22 \cite{zerva2022findings}. WMT22 results are averaged over $\textrm{x}\rightarrow\textrm{en}$ translations for different languages $\textrm{x}$. For TydiQA, we pass the gold passage as part of the input, and report the average F1-score over all languages. For SuperGLUE and Gen-tasks, we follow the experimental settings as described in \cite{palm2} and report the average results. We report 1-shot evaluations for all performance evaluation experiments.

\vspace{2.5mm}

\subsection{Latency Evaluation}
We perform latency evaluation in two different settings. In the first setting, we use \Tandemdistill as the secondary model within SPEED, with \bison as the primary model. Note that the SPEED framework guarantees that the outputs will be of the same quality as the primary model \bison. For comparison, we use \bison as a stand alone model, as well as SPEED with \bison as the primary model and \geckodistill as the secondary model as our baselines. In the second setting, we evaluate the latency of \Tandemce and \Tandemdistill as stand alone models with \gecko, \otter and \bison. All the evaluations are performed on TPUv5e~\cite{tpuv5e}.
% our Tandem model, and PaLM2-Gecko-Distil are combined with the SPEEDframework to guarantee PaLM-Bison's performance.

We evaluate latency on the test sets of CNN/DailyMail \cite{hermann2015teaching}, Reddit Posts summarization \cite{DBLP:journals/corr/abs-1811-00783}, and 1000 prompts from the 1 Billion Word Benchmark \cite{DBLP:conf/interspeech/ChelbaMSGBKR14}. We report latency results for both $\numsamples = 1$ as well as $4$.

\subsection{Evaluation Results}
% In this section, we train only the secondary model $\Ms$ in the Tandem setup, while freezing the primary model $\Ml$. For the Tandem model, we initialize the secondary model $\Ms$ to be the pretrained PaLM2-Gecko, and the primary model $\Ml$ to be the pretrained PaLM2-Bison~\cite{palm2}. Our baseline, PaLM2-Gecko-Distil, is obtained through logit distillation between PaLM2-Gecko and PaLM2-Bison. All distillation experiments use a temperature $\tau =1$ for the drafter $Ms$ and a temperature $\tau =0.4$ for the primary model $Ml$. Also, our Tandem model is trained with a block length of $\gamma = 2$.
We now present results of our evaluation of Tandem Transformers.

{\bf Pretraining Metrics.} Table~\ref{tab:flat-pretraining} presents a comparison of accuracy and cross entropy (CE) loss of various baselines as well as Tandem models, with respect to both the ground truth labels as well as the primary model $\Ml$'s predictions. As we can see, Tandem Transformers performs better than logit distillation, while combining logit distillation with Tandem Transformers, further improves its performance.

\vspace{1.5mm}
{\bf Latency within SPEED.}
Table~\ref{tab:flat-latency} presents results on the latency of Tandem Transformers within the SPEED framework. Specifically, we compare the speedup obtained over the \bison model, by using SPEED with \geckodistill as the secondary model vs Tandem-Distil as the secondary model. The results clearly demonstrate the improvements obtained by Tandem on top of distillation. Table~\ref{tab:decode-latency} in Appendix~\ref{app:additional-results} presents the speedups computed only over the decode time (i.e., excluding the query processing time). Note that since the SPEED framework guarantees that the outputs are of same quality as those of the primary model, \bison, the latency improvements given by the Tandem model do not have any quality tradeoffs.
% \begin{table*}[t]
% \begin{center}
% \resizebox{\textwidth}{!}{
% \begin{tabular}{|c|c|c|c|c|c|c|}
%  \hline
%  \multirow{2}{*}{Dataset} & \multirow{2}{*}{Num samples} & \multirow{2}{*}{$\gamma^{\text{optimal}}_{\text{PaLM-Gecko-Distil + SPEED}}$} & \multirow{2}{*}{$\gamma^{\text{optimal}}_{\text{Tandem + SPEED}}$} & PaLM-Gecko-Distil + SPEED's  & Tandem + SPEED's  & Tandem + SPEED's speedup \\ & & & & speedup over PaLM-Bison & speedup over PaLM-Bison & over PaLM-Gecko-Distil + SPEED \\ 
%  \hline \hline
%  \multirow{2}{*}{\centering Reddit} & 1&5 & 7&	$2.105 \times$ &	$2.471 \times$& $1.174 \times$ \\
%  \cline{2-7}
%  & 4& 5 & 7& $1.862 \times$ & $2.234\times$ & $1.2 \times$	\\
%  \hline\hline
% \multirow{2}{*}{\centering CNN/DailyMail} & 1&7 &7 &	$2.145 \times$&	$2.473 \times$ & $1.153 \times$ \\
%  \cline{2-7}
%  & 4& 5 & 7& $1.877 \times$ & $2.19 \times$ & $1.167 \times$\\
%  \hline\hline
% \multirow{2}{*}{\centering LM1B} & 1& 7& 7&	$2.283 \times$ &	$2.61 \times$& $1.143 \times$  \\
%  \cline{2-7}
%  & 4& 5 &7 & $1.952 \times$& $2.359 \times$& $1.208 \times$	\\
%  \hline
% \end{tabular}
% }
% \end{center}
% \caption{End-to-end latency when used within the SPEED framework with \bison as the primary model, for various choices of secondary models. Here, $\gamma^{\text{optimal}}_{\text{PaLM-Gecko-Distil + SPEED}}$ is obtained by optimizing for PaLM-Gecko-Distil + SPEED's latency. Similarly $\gamma^{\text{optimal}}_{\text{Tandem + SPEED}}$ is obtained by optimizing for Tandem + SPEED's latency. \pn{TODO.}}
% \label{tab:flat-latency}
% \end{table*}

\begin{table*}[htbp!]
\caption{End-to-end latency gain of various secondary models, when used within the SPEED framework with \bison as the primary model. The secondary models we consider are: \xxxsdistill and Tandem-Distil (with \xxxs as the secondary model). For each secondary model, we tune the optimal block length $\gamma$ parameter. We consider two settings, one where we generate a single response and another where we generate $4$ responses for the given query. The third and fourth column provide the speedup by using \xxxsdistill and Tandem models respectively, with respect to the \bison model. The last column indicates the relative gain of using the Tandem model as the secondary model in SPEED, instead of \xxxsdistill. The results clearly demonstrate the additional improvements Tandem obtains, on top of logit distillation. Note that both \xxxsdistill and Tandem-Distil (with \xxxs as the secondary model) improve over their \gecko counterparts.}
\vspace{2mm}
% \begin{center}
\centering
% \resizebox{\textwidth}{!}{
\begin{tabular}{lcccc}
 \toprule
 \multirow{2}{*}{Dataset} & num- & PaLM2-XXXS-Distil & Tandem-Distil  & Tandem-Distil \\ & samples & (baseline) & (\textbf{ours}) & (\textbf{ours}; relative gain) \\ 
 \midrule
\multirow{2}{*}{\centering LM1B} & $1$& $2.445 \times$ \text{   
 }($\gamma=5$)&	$3.040 \times$ \text{   
 }($\gamma=5$)& $\mathbf{1.243} \times$  \\
 % \cline{2-5}
 & $4$& $ 1.182 \times$ \text{  
 }($\gamma=5$)& $2.488 \times$ \text{   
 }($\gamma=5$)& $\mathbf{1.366} \times$	\\
 \bottomrule
\end{tabular}
% }
% \end{center}
\label{tab:xxxs-latency}
\end{table*}

\begin{table}[t!]
\caption{End-to-end latency speedup obtained by Tandem-Distil + SPEED + Adaptive $\gamma$ on different evaluation datasets. The second and third columns show the speedup over the stand alone \bison model and Tandem-Distil + SPEED model respectively. The latency is evaluated for generating a single response. Adaptive $\gamma$ enables us to use much larger block lengths without losing performance. For example, on the Reddit dataset, the optimal $\gamma$ for the Tandem model in the standard SPEED setup is $7$, while adaptive $\gamma$ obtains better results with $\gamma_{\textrm{max}}=17$.}
\vspace{2mm}
\centering
\resizebox{\columnwidth}{!}{
\begin{tabular}{lcc}
\toprule
 {Dataset} & PaLM-Bison & Tandem-Distil + SPEED \\  
 \midrule
 Reddit & $2.582 \times$ \text{   
 }($\gamma_{\text{max}}=17$)& $\mathbf{1.045} \times$	\\ 
 % \midrule
CNN/ & \multirow{2}{*}{$2.599 \times$}	\multirow{2}{*}{\text{   
 }($\gamma_{\text{max}}=17$)} &  \multirow{2}{*}{$\mathbf{1.051} \times$} \\
DailyMail & & \\
 % \midrule
 LM1B & $2.853 \times$ \text{   
 }($\gamma_{\text{max}}=27$)& $\mathbf{1.093} \times$	\\
 \bottomrule
\end{tabular}
}
\label{tab:flat-latency-AG}
\end{table}
\begin{table}[h]
% \begin{center}
\caption{Primary model and secondary model runs for Tandem-Distil and Tandem-Distil + AG on the LM1B benchmark. Note that these results are obtained for $\numsamples = 1$. We can see that the number of secondary model runs have come down by $90$ whereas the number of large model runs has gone up only by $3$. The results clearly showcase that an adaptive block length can significantly cut down on the number of secondary model runs and give non-trivial latency gains.}
\centering
\vspace{2mm}
\resizebox{\columnwidth}{!}{
\begin{tabular}{lcc}
\toprule
   &  Tandem-Distil & Tandem-Distil + AG \\ 
 \midrule
 Primary model runs & 51.53 & 54.67 \\ 
 % \midrule
 Secondary model runs & 360.73 & 271.63 \\ 
 \bottomrule
\end{tabular}
}
% \end{center}
\label{tab:AG-gains}
\end{table}

\vspace{2mm}
{\bf Evaluation as a Standalone Model.}
We evaluate the Tandem model as a stand alone model in its own right.
Table~\ref{tab:flat-downstream} presents a comparison of both downstream evaluations on standard downstream benchmarks, as well as latency evaluations. As can be seen, the Tandem model substantially improves upon the downstream performance of the baseline model, and is almost on par with the \otter model. Detailed results presented in Tables~\ref{tab:flat-downstream-gen-tasks} and~\ref{tab:flat-downstream-superglue} in Appendix~\ref{app:additional-results} show that, in some cases, the Tandem model is closer to the \bison model itself. At the same time, the Tandem model is about $1.16 \times$ times faster compared to the \otter model, making it a compelling candidate for stand alone deployment as well.

% \subsection{primary model $\Ml$ allowed to train}
% In this section, we allow both $\Ml$ and $\Ms$ to train from pretrained checkpoints.
% For the Tandem model, we initialize the secondary model $\Ms$ to be the pretrained PaLM2-Otter, and the primary model $\Ml$ to be the pretrained PaLM2-Bison~\cite{palm2}.
% Similar to the previous section, we again evaluate the performance of the Tandem model on three axes: pretraining metrics, downstream metrics and inference latency. Since the primary model $\Ml$ changes compared to the pretrained version, the inference latency evaluations will be for a stand along Tandem model, without using the SPEED framework.
% \praneeth{TODO
% \begin{itemize}
%     \item Pretraining metrics
%     \item Downstream metrics
%     \item Latency
% \end{itemize}
% }
% \subsubsection{Illustrative examples of generation}
% Improvement on head tasks.
{\bf Adaptive Block Length.} We now present a way to improve the performance of SPEED with adaptive block lengths (Adaptive $\gamma$ or AG), where after every token predicted by the secondary model, we use a small, inexpensive router to determine whether to continue predicting with the secondary model, or verify the tokens generated so far with the primary model. Table~\ref{tab:flat-latency-AG} presents the speedup obtained by Tandem-Distil + SPEED + AG compared with the \bison model as well as the Tandem-Distil + SPEED model. Table~\ref{tab:decode-latency-AG} in Appendix~\ref{app:additional-results} presents the speedup as measured only over the decode component of the latency i.e., excluding query processing time.

In Table~\ref{tab:AG-gains}, we present the number of primary model, and secondary model runs for Tandem-Distil + SPEED and Tandem-Distil + SPEED + Adaptive $\gamma$. The results put forth the benefits of using an adaptive block length, since it drastically reduces the number of secondary model runs while slightly increasing the number of primary model runs.

{\bf Smaller Secondary Model.}
% While both PaLM2-Gecko-Distil and Tandem-Distil show significant improvements over PaLM2-Bison, we find that using a smaller secondary model further boosts the latency improvements. Table~\ref{tab:xxxs-latency} presents results for the same where \gecko is replaced with \xxxs, with \bison being retained as the primary model. As can be seen, this setting further improves the latency gains of the Tandem model, indicating that Tandem Transformers provide considerably better latency, even with extremely small secondary models.
\vspace{-1mm}
While Tandem-Distil show significant improvements over PaLM2-Bison, we find that using a smaller secondary model for Tandem Transformers further boosts the latency improvements. Table~\ref{tab:xxxs-latency} presents results for the same where \gecko is replaced with \xxxs, with \bison being retained as the primary model. As can be seen, this setting further improves the latency gains of the Tandem model, whereas PaLM2-XXXS-Distil improves over its \gecko counterpart only for $\numsamples = 1$, indicating that Tandem Transformers provide considerably better latency, even with extremely small secondary models.

\vspace{-2mm}
\section{Deep Tandem Transformers}
\begin{figure}
    \centering
    % \hspace{-20mm}
    \includegraphics[scale=0.5]{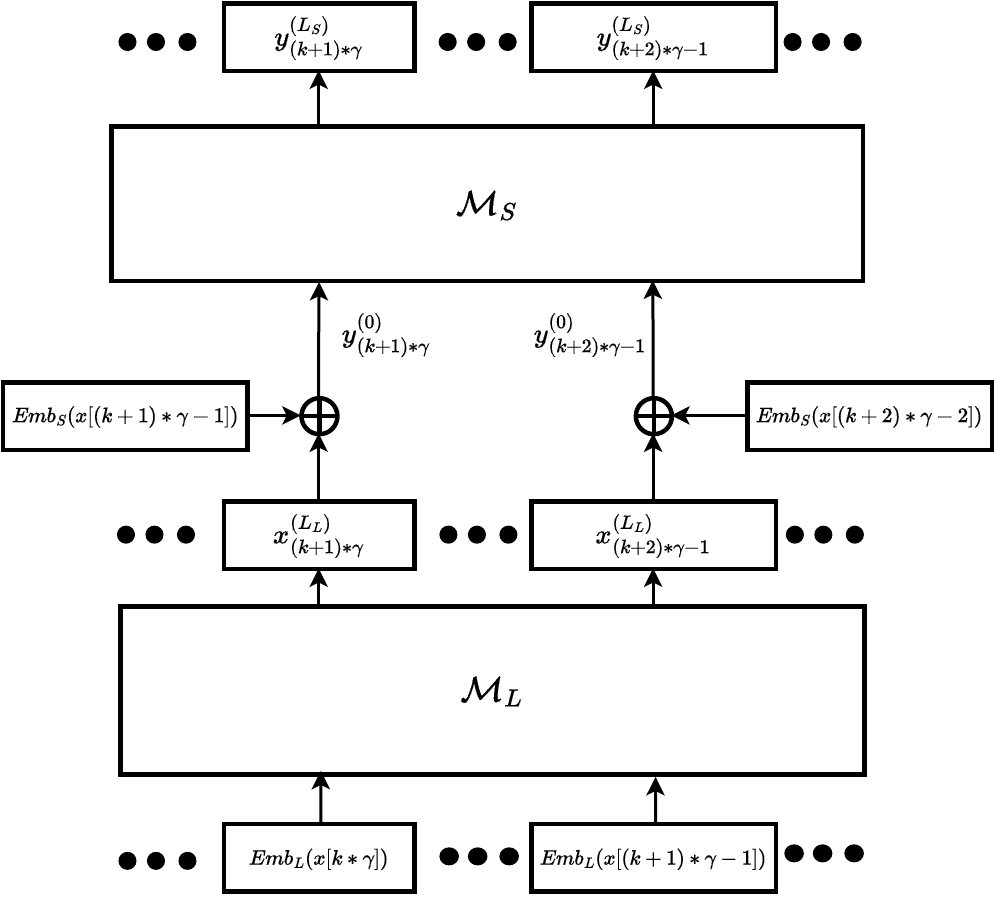}
    % \vspace{-12mm}
    
    \caption{The architecture of Deep Tandem Transformers with a block length of $\gamma$. See text and Equations~\eqref{eqn:deep-Tandem-1} and~\eqref{eqn:deep-Tandem-2} for description.}
    \label{fig:deep-Tandem}
\end{figure}

In Tandem Transformers, we used the large model $\Ml$ to process tokens in blocks, so that the small model $\Ms$ can use large model's representations for all the tokens from previous blocks. In this section, we present a different approach to use $\Ml$ and $\Ms$ in Tandem, where $\Ml$ predicts a sketch of the next block of tokens in parallel, while $\Ms$ does the actual sampling in an autoregressive manner. More concretely, we have:
\begin{align}
    \xtilde{i}{j+1} &= \attnl{\xss{i}{j}}{\xss{\leq k*\gamma}{j}}{j+1} \mbox{ where } k = \ceil{\frac{i-\gamma}{\gamma}} \nonumber \\
    \xss{i}{j+1} &= \ffnl{\xtilde{i}{j+1}}{j+1} \qquad \mbox{ for } j = 0,\cdots,L_L-1, \label{eqn:deep-Tandem-1}
\end{align}
and $\xss{i}{0}=\embl{\xindex{{i-\gamma}}}$ is given by the large model's embedding of the $\left(\ceil{\frac{i-\gamma}{\gamma}}*\gamma\right)^\textrm{th}$ token, where the large model, given all tokens $\xss{1}{0},\cdots,\xss{s*\gamma}{0}$, produces a draft of the next $\gamma$ tokens $\xss{k*\gamma+1}{L_L},\cdots,\xss{(k+1)*\gamma}{L_L}$. We then add the previous token representations to these sketches and then pass it through the small model, which predicts the next token autoregressively:
\begin{align}
    \yss{i}{0} &= \embs{\xindex{{{i-1}}}} + \ffndeepTandem{\xss{i}{L_L}} \nonumber\\
    \ytilde{i}{j+1} &= \attns{\yss{i}{j}}{\yss{\leq i}{j}}{j+1} \nonumber \\
    \yss{i}{j+1} &= \ffns{\ytilde{i}{j+1}}{j+1} \qquad \mbox{ for } j = 0,\cdots,L_S-1. \label{eqn:deep-Tandem-2}
\end{align}
The eventual output of the model is $\yss{i}{L_S}$ which is its prediction of the $i^\textrm{th}$ token in the input sequence. This is pictorially depicted in Figure~\ref{fig:deep-Tandem}.

\section{Conclusions and Discussion}\label{sec:conc}
In this work, we introduce a novel architecture, Tandem Transformers, which combines a small autoregressive model with a large model operating in block mode. Tandem Transformers substantially boost the small model's predictive accuracy by allowing it to attend to representations from the large model. In our experiments, a Tandem model comprising of \bison and \gecko substantially improves over a standalone \gecko, and gives comparable performance to the \otter model, while being $1.16 \times$ faster than the \otter model. When used within the SPEED setup as a secondary model, the distilled \emph{Tandem} \gecko model gives around $1.14 \times$ speedup over a distilled \gecko model. We further improve our Tandem model through an adaptive block length procedure in SPEED and obtain around $1.22 \times$ speedup over using \geckodistill as the secondary model.

\paragraph{Limitations and Future directions}
\begin{itemize}
    % \item \textbf{Other variants of Tandem}: In our current approach, we use the large model only through its representations of the past tokens. Is it possible to use the large model to also generate a \emph{plan for the future $\gamma$ tokens}?
    % Allow for non-auto-regressive generation through plan generation by the large model.
    \item \textbf{Alternative to LoRA for finetuning}: The current approach for finetuning a base model for multiple downstream applications is through low rank adaptation (LoRA)~\cite{hu2021lora}. It will be interesting to explore whether Tandem with block length $0$ can be an effective alternative to LoRA, while reducing the training cost substantially since backpropagation needs to be done only for the small model.
    \item \textbf{Adaptive $\gamma$ for larger \numsamples /batch-size:} While we see promising results with adaptive $\gamma$ in SPEED for num samples $1$, extending it to larger num samples seems challenging. Identifying an effective way of determining when to continue generating with small model vs verifying with large model, in the larger num samples setting, is also an interesting direction of future work.
    % \item \textbf{Smaller drafter models in SPEED}: Finally, we hope that Tandem can enable using even smaller drafter models in SPEED, compared to the ones currently being pursued, leading to both memory as well as latency improvements.
    % Our current implementation of adaptive $\gamma$ for num-samples $>$ 1 involves continuing generation with the secondary model until and unless all the samples want to fall back to the primary model. However, by doing so, we are generating several unnecessary tokens for some sample that wants to exit early. This forces us to use a small $\gamma_{max}$, thus significantly reducing our gains. In future, we would like to refine our exit strategy to get latency gains for num-samples $>$ 1. 
\end{itemize}

% \input{disc-deepTandem}

% \section*{Acknowledgements}

\section*{Impact Statement}
Our work provides a more computationally efficient large language model inference solution, which we hope can bring down carbon emissions associated with LLM inference. It also helps with easier deployment of LLMs, which could have potential societal consequences, that seem difficult to predict. 

\bibliography{main_icml}
\bibliographystyle{icml2024}

%%%%%%%%%%%%%%%%%%%%%%%%%%%%%%%%%%%%%%%%%%%%%%%%%%%%%%%%%%%%%%%%%%%%%%%%%%%%%%%
%%%%%%%%%%%%%%%%%%%%%%%%%%%%%%%%%%%%%%%%%%%%%%%%%%%%%%%%%%%%%%%%%%%%%%%%%%%%%%%
% APPENDIX
%%%%%%%%%%%%%%%%%%%%%%%%%%%%%%%%%%%%%%%%%%%%%%%%%%%%%%%%%%%%%%%%%%%%%%%%%%%%%%%
%%%%%%%%%%%%%%%%%%%%%%%%%%%%%%%%%%%%%%%%%%%%%%%%%%%%%%%%%%%%%%%%%%%%%%%%%%%%%%%
\newpage
\appendix
\onecolumn
\section{Additional Results}\label{app:additional-results}
In this section, we will present additional experimental results.

\subsection{Decode Time Results}
In Tables~\ref{tab:decode-latency} and~\ref{tab:decode-latency-AG}, we compare the decode time results ($\text{i.e. end-to-end time}  - \text{time required to process the input prefix}$) of our Tandem model and its Adaptive $\gamma$ variant with the baselines.

\begin{table*}[h]
\caption{Decode-time-only latency gain of various secondary models, when used within the SPEED framework with \bison as the primary model. The secondary models we consider are: \geckodistill and Tandem-Distil. For each secondary model, and on each dataset, we use the optimal block length $\gamma$ parameter. We consider two settings, one where we generate a single response and another where we generate $4$ responses for the given query. The third and fourth column provide the speedup by using \geckodistill and Tandem models respectively, with respect to the \bison model. The last column indicates the relative gain of using the Tandem model as the secondary model in SPEED, instead of \geckodistill. The results clearly demonstrate the additional improvements Tandem obtains, on top of logit distillation.}
% \begin{center}
% \resizebox{\textwidth}{!}{
\centering
\vspace{2mm}
\begin{tabular}{lcccc}
 \toprule
 \multirow{2}{*}{Dataset} & {num-} & PaLM2-Gecko-Distil & Tandem-Distil  & Tandem-Distil \\ & samples & (baseline) & (\textbf{ours}) & (\textbf{ours}; relative gain) \\ 
 \midrule
 \multirow{2}{*}{\centering Reddit} & 1&	$2.356 \times$ \text{   
 }($\gamma=7$) &	$2.737 \times$ \text{   
 }($\gamma=7$) & $\mathbf{1.162							} \times$ \\ 
 % \cline{2-5}
 & 4& $2.042   \times$ \text{ 
 }($\gamma=5$)& $2.425\times$ \text{   
 }($\gamma=7$)& $\mathbf{1.188						} \times$	\\
 \midrule
\multirow{2}{*}{\centering CNN/DailyMail} & 1&	$2.418 \times$ \text{   
 }($\gamma=7$)&	$2.740 \times$ \text{   
 }($\gamma=7$)& $\mathbf{1.133							} \times$ \\
 % \cline{2-5}
 & 4& $2.066 \times$ \text{   
 }($\gamma=5$)& $2.369 \times$ \text{   
 }($\gamma=7$)& $\mathbf{1.146						} \times$\\
 \midrule
\multirow{2}{*}{\centering LM1B} & 1& $2.460 \times$ \text{   
 }($\gamma=7$)&	$2.756 \times$ \text{   
 }($\gamma=7$)& $\mathbf{1.120} \times$  \\
 % \cline{2-5}
 & 4& $ 2.080 \times$ \text{  
 }($\gamma=5$)& $2.466 \times$ \text{   
 }($\gamma=7$)& $\mathbf{1.186						} \times$	\\
 \bottomrule
\end{tabular}
% }
% \end{center}
\label{tab:decode-latency}
\end{table*}

\begin{table}[h]
\caption{Decode-time-only latency speedup obtained by Tandem-Distil + SPEED + Adaptive $\gamma$ on different evaluation datasets. The second and third columns show the speedup over the stand alone \bison model and Tandem-Distil + SPEED model respectively. The latency is evaluated for generating a single response. Adaptive $\gamma$ enables us to use much larger block lengths without losing performance. For example, on the Reddit dataset, the optimal $\gamma$ for the Tandem model in the standard SPEED setup is $7$, while adaptive $\gamma$ obtains better results with $\gamma_{\textrm{max}}=17$.}
\centering
\vspace{2mm}
% \resizebox{\textwidth}{!}{
\begin{tabular}{lcc}
\toprule
 {Dataset} & PaLM-Bison & Tandem-Distil + SPEED \\  
 \midrule
 Reddit & $2.885 \times$ \text{   
 }($\gamma_{\text{max}}=17$)& $\mathbf{1.054} \times$	\\ 
 % \midrule
CNN/DailyMail & {$2.908 \times$}	{\text{   
 }($\gamma_{\text{max}}=17$)} &  {$\mathbf{1.061} \times$} \\
% DailyMail & & \\
 % \midrule
 LM1B & $3.040 \times$ \text{   
 }($\gamma_{\text{max}}=27$)& $\mathbf{1.103} \times$	\\
 \bottomrule
\end{tabular}
% }
\label{tab:decode-latency-AG}
\end{table}

\begin{table}[h]
\caption{Evaluation of the Tandem model on each of the Generative-tasks. We see that the Tandem model substantially improves upon the performance of stand alone \gecko model, and on most datasets, is on par with the \otter model. On the other hand, the latency evaluations in the last row demonstrate that the Tandem model is about $1.16 \times$ faster than the \otter model.}
% \begin{center}
\centering
\vspace{2mm}
\begin{tabular}{lccccc}
 \toprule
 \multirow{2}{*}{Dataset} & \gecko & Tandem-CE & Tandem-Distil  & \otter & \bison \\ 
 & & \textbf{(ours)} & \textbf{(ours)} & & \\
 \midrule
 Lambada (acc = Accuracy) & $45.5$ & $59.2$ & $68.3$ & $78.9$ & $82.9$\\
 % \midrule
 % GPT3-Rank & 57.1 & 70.3 & 70.2 & & 73.6\\ [1ex]
 % \hline
 NaturalQuestions (em = Exact Match) & $7.7$ & $9.9$ & $14.4$ & $19.9$ & $28.1$\\ 
 % \midrule
 SQuADv2 (em) & $45.3$ & $67.8$ & $70.2$ & $70.3$ & $75.4$ \\ 
 % \midrule
 TriviaQA (em) & $36.8$ & $36.9$ & $51.2$ & $68.9$ & $77.3$ \\ 
 % \midrule
 WebQuestions (em) & $9.0$ & $12.0$ & $16.0$ & $17.6$ & $23.8$ \\ 
 \bottomrule
\end{tabular}
% \end{center}
\label{tab:flat-downstream-gen-tasks}
\end{table}

\begin{table}[h]
\caption{Evaluation of the Tandem model on each of the SuperGLUE tasks. We see that the Tandem model substantially improves upon the performance of stand alone \gecko model, and on most datasets, is on par with the \otter model. On the other hand, the latency evaluations in the last row demonstrate that the Tandem model is about $1.16 \times$ faster than the \otter model.}
% \begin{center}
\centering
\vspace{2mm}
\begin{tabular}{lccccc}
 \toprule
 \multirow{2}{*}{Dataset} & \gecko & Tandem-CE & Tandem-Distil  & \otter & \bison \\ 
 & & \textbf{(ours)} & \textbf{(ours)} & & \\
 \midrule
 BoolQ (acc) & $65.4$ & $87.8$ & $87.6$ & $85.5$ & $88.8$\\
 % \midrule
 % GPT3-Rank & 57.1 & 70.3 & 70.2 & & 73.6\\ [1ex]
 % \hline
 CB (acc) & $39.3$ & $82.1$ &  $83.9$ & $71.4$ & $87.5$\\ 
 % \midrule
 COPA (acc) & $80.0$ & $78.0$ & $82.0$ & $88.0$ & $88.0$ \\ 
 % \midrule
 RTE (acc) & $55.2$ & $80.1$ & $78.3$ & $84.1$ & $77.6$\\ 
% \midrule
 ReCoRD (acc) & $85.5$ & $87.8$ & $87.2$ & $91.2$ & $92.2$\\ 
 % \midrule
 WIC (acc) & $47.5$ & $50.0$ & $50.6$ & $49.7$ & $50.9$ \\ 
% \midrule
 WSC (acc) & $75.8$ & $81.1$ & $80.4$ & $86.3$ & $86.3$\\ 
 % \midrule
MultiRC (F1) & $53.9$ & $80.8$ & $80.1$ &  $76.1$ & $80.5$\\ 
 \bottomrule
\end{tabular}
% \end{center}
\label{tab:flat-downstream-superglue}
\end{table}

\subsection{Detailed Performance Evaluation Results}
In Table~\ref{tab:flat-downstream-gen-tasks}, we present results for our Tandem model and the compared baselines on each individual task in Generative-tasks. Likewise, in Table~\ref{tab:flat-downstream-superglue} we present results on each individual task in SuperGLUE.

\section{Inference of Tandem Transformers}\label{no-token}
\begin{figure*}[t!]
\centering
% {0.45\textwidth}
% \begin{minipage}
%     \centering
    \includegraphics[width=1\textwidth]{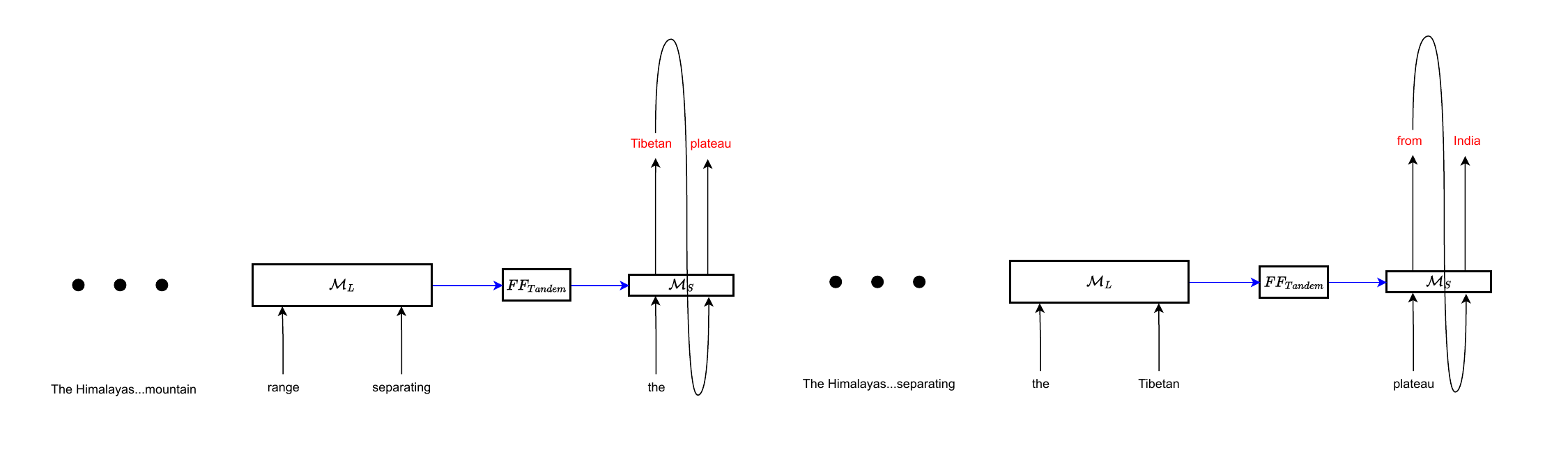}
% \end{minipage}
% \begin{minipage}{0.45\textwidth}
%     \centering
%     \includegraphics[width=1\textwidth]{figures/TandemInference2_nofreetoken.drawio.pdf}
% \end{minipage}
\vspace{-12mm}
    \caption{Inference of Tandem Transformers \emph{without} free token from the primary model $\Ml$. (left) First block prediction. (right) Second block prediction. Given the same query \emph{The Himalayas are a mountain range separating the} as in Figure~\ref{fig:inference-free-token}, here, $\Ml$ first processes this query except the last token \emph{the}. The last token is passed as an input to the secondary model $\Ms$, which attends to $\Ml$ representations for all past tokens, and produces the first block of responses \red{Tibetan plateau} autoregressively. In the second block, $\Ml$ processes \emph{the Tibetan} in a block mode while \emph{plateau} is passed as an input to $\Ms$, which then autoregressively generate the next block of response \red{from India}. This eventually leads to a response of \red{Tibetan plateau from India...}.}
    \label{fig:inference-no-free-token}
\end{figure*}

Figure~\ref{fig:inference-no-free-token} presents the inference for Tandem Transformers without the the free token from the primary model $\Ml$.

% \clearpage

% \section{You \emph{can} have an appendix here.}

% You can have as much text here as you want. The main body must be at most $8$ pages long.
% For the final version, one more page can be added.
% If you want, you can use an appendix like this one.  

% The $\mathtt{\backslash onecolumn}$ command above can be kept in place if you prefer a one-column appendix, or can be removed if you prefer a two-column appendix.  Apart from this possible change, the style (font size, spacing, margins, page numbering, etc.) should be kept the same as the main body.
%%%%%%%%%%%%%%%%%%%%%%%%%%%%%%%%%%%%%%%%%%%%%%%%%%%%%%%%%%%%%%%%%%%%%%%%%%%%%%%
%%%%%%%%%%%%%%%%%%%%%%%%%%%%%%%%%%%%%%%%%%%%%%%%%%%%%%%%%%%%%%%%%%%%%%%%%%%%%%%

\end{document}